\documentclass[nohyperref]{article}

\usepackage{microtype}
\usepackage{graphicx}
\usepackage{subfigure}
\usepackage{booktabs} 

\usepackage{hyperref}
\usepackage{fancyhdr}

 \usepackage[accepted]{icml2022}

\usepackage{amsmath}
\usepackage{amssymb}
\usepackage{mathtools}
\usepackage{amsthm}
\usepackage{multirow}
\usepackage[capitalize,noabbrev]{cleveref}

\theoremstyle{plain}

\theoremstyle{definition}

\theoremstyle{remark}

\newcommand{\mL}{\mathcal{L}}
\newcommand{\mR}{\mathbb{R}}
\newcommand{\bx}{\mathbf{x}}
\newcommand{\by}{\mathbf{y}}
\newcommand{\bz}{\mathbf{z}}
\usepackage[textsize=tiny]{todonotes}

\icmltitlerunning{Submission and Formatting Instructions for ICML 2022}

\begin{document}

\twocolumn[
\icmltitle{Attention-based Multi-task Learning for Base Editor Outcome Prediction}



\icmlsetsymbol{equal}{*}

\begin{icmlauthorlist}
\icmlauthor{Amina Mollaysa}{equal,yyy,athai}
\icmlauthor{Ahmed Allam}{equal,yyy}
\icmlauthor{Michael Krauthammer}{yyy,athai}
\end{icmlauthorlist}

\icmlaffiliation{yyy}{Department of Quantitative Biomedicine, University of Zurich, Zurich, Switzerland}
\icmlaffiliation{athai}{ETH AI Center, Zurich, Switzerland}

\icmlcorrespondingauthor{Amina mollaysa}{maolaaisha.aminanmu@uzh.ch}

\icmlkeywords{Machine Learning, ICML}

\vskip 0.3in
]
\printAffiliationsAndNotice{\icmlEqualContribution} 
\begin{abstract}
Human genetic diseases often arise from point mutations, emphasizing the critical need for precise genome editing techniques. Among these, base editing stands out as it allows targeted alterations at the single nucleotide level. However, its clinical application is hindered by low editing efficiency and unintended mutations, necessitating extensive trial-and-error experimentation in the laboratory.  To speed up this process, we present an attention-based two-stage machine learning model that learns to predict the likelihood of all possible editing outcomes for a given genomic target sequence.  We further propose a multi-task learning schema to jointly learn multiple base editors (i.e. variants) at once. Our model's predictions consistently demonstrated a strong correlation with the actual experimental results on multiple datasets and base editor variants. These results provide further validation for the models' capacity to enhance and accelerate the process of refining base editing designs. 
\end{abstract}

\section{Introduction}
The landscape of human genetic diseases is largely dominated by a significant proportion of cases arising from point mutations \citep{landrum2014clinvar}. These minor genetic alterations (substitution, deletion, or insertion of a single nucleotide), pose serious implications for health ranging from rare monogenic conditions to more common chronic diseases such as obesity, coronary heart disease, diabetes, Alzheimer’s disease, and multiple cancers.
Genome editing approaches allow researchers to make specific/targeted changes to the DNA of living cells. Among these, base editing
\citep{komor2016programmable, gaudelli2017programmable, rees2018base} shows promising results as it enables precise genome editing at the single nucleotide level without causing double-stranded breaks (DSBs) in the DNA. Base editors (BE) operate by combining catalytically impaired Cas nucleases of the CRISPR-Cas system \footnote{See appendix section \ref{crispr}} 
with enzymes capable of converting one DNA base pair into another. 
While base editors have great potential as genome editing tools for basic research and gene therapy, their application has been limited due to 1) low editing efficiency on specific sequences or 2) unintended editing results with concurrent mutations, especially where there are multiple substrate nucleotides within close proximity to the intended edit.

The development of a robust machine learning model capable of accurately predicting the potential editing outcomes of diverse base editors on various target sites (i.e. genomic sequences) could significantly enhance the field. By using such models, biologists would assess possible outcomes much faster and fine-tune their editing strategies with high efficiency. Moreover, these models would help in exploring the different factors affecting the editing efficiency of the various base editors, offering a great aid for knowledge discovery.

In this paper, we focus on predicting the potential outcome sequences when various base editors are applied to specific DNA targets. Given a target sequence (i.e. reference genomic sequence), the goal is to learn the probability distribution of all outcome sequences (i.e. sequences resulting from different edits of a reference sequence) when applying a given base editor (Figure \ref{fig:data_demo}). In literature, the reference sequence that remains the same after applying a base editor is denoted by a 
\emph{wild-type} sequence. Furthermore, edits are defined by a change of letters (i.e. nucleotides) in a reference sequence.

Rather than using a sequence-to-sequence model \citep{sutskever2014sequence} that generates complete output sequences, we leverage the inherent characteristics of base editing, which result in minimal positional alterations, generating edited sequences that differ from the reference sequence by only a few positions. We consider all the possible outcomes prior to the model and learn the likelihood of an outcome sequence by considering both the target DNA sequence and the outcome sequence.

We explore two modeling ideas. First is a one-stage model where we directly learn the probability distribution over all possible outcome sequences for a given target sequence. The second one is a two-stage model where we first estimate the probability of the given target sequence being edited, acknowledging that in many cases, the editor fails and no changes are observed which is often referred to as wild-type outcome. We then proceed to estimate the probability distribution of edited outcomes. 



Different editors exhibit varying behaviors on the same target sequences due to factors like binding affinities and editing window sizes, introducing distributional shifts. In response to this challenge, we introduce a multi-task learning framework. Rather than training individual models for each editor, as current models do, we propose a unified model capable of simultaneously accommodating multiple editors.


In this work, we study the different modeling strategies for training machine learning models for the base editor outcome prediction task. We explore the spectrum of modeling choices evaluated on multiple datasets and base editors. A key highlight is the proposed unified multi-task model that is capable of learning from various base editors without necessitating training separate models for each setup.  We train our models on six libraries corresponding to the outcomes of six base editors applied on thousands of target sites (Table \ref{tab:reference_seq}). Our models' predictions show a good correlation with the ground truth across all datasets demonstrating the potential of machine learning in guiding and exploring genome editing space.

\section{Related Work}
In recent years, the intersection of deep learning and CRISPR-Cas9 systems has witnessed substantial interest from the bioinformatics community. Researchers have explored the applications of deep learning in predicting various aspects of CRISPR-Cas9 systems, including predicting gRNA activities \citep{ameen2021c, xie2023crispr, zhang2021prediction} and editing outcomes for both base editing and prime editing scenarios \citep{mathis2023predicting}.

Among those, one notable approach is the BE-Hive proposed by \cite{arbab2020determinants}, which aims to predict base editing outcomes and efficiencies while considering sequence context, PAM compatibility, and cell-type-specific factors. The model employs a gradient boosting tree for predicting overall editing efficiency and a deep conditional autoregressive model for predicting probability of edited outcome sequences (denoted by bystander efficiency). Similarly, \cite{song2020sequence} presented DeepABE and DeepCBE, that is based on convolutional neural networks to model both overall editing efficiency and bystander efficiency of adenine and cytosine base editors. 

Recently, \citet{marquart2021predicting} proposed BE-DICT, which predicts per-base editing efficiency (i.e. editing efficiency of each target base in a sequence) and bystander base-editing efficiency using attention-based deep learning. 
In a latest comprehensive study, \cite{kim2023deep} developed DeepCas9variants and DeepBEs to predict editing efficiencies and outcomes of various BEs, taking into account different Cas9 variants. They build on and adapt the models proposed in \cite{song2020sequence} (i.e. convolutional networks) to generate predictions for a range of CRISPR-Cas9 systems.

While the surge of interest in applying machine learning to CRISPR-Cas9 systems is clear in recent literature, it's noteworthy that many of these works have a primary emphasis on designing CRISPR-Cas9 systems under various conditions and less focused on the analysis of ML models without offering a holistic and systematic analysis of model design. Given the intricate nature of CRISPR-Cas9 systems and the multitude of model paradigms adopted, deriving concrete conclusions about optimal model design strategies remains elusive. In this context, our work aims to serve as model-first work that presents the base editing outcome prediction through a modeling lens. We focus on model development and provide a systematic analysis of each component of the models, providing a structured framework for problem formulation and model design specifically tailored to the prediction of base editing outcomes.  Through this structured examination of these critical aspects, our aim is to lay the groundwork for more informed and refined approaches for using deep learning models to assist the design of base editors.


\section{Method}
\begin{figure}[tb]
    \centering
    \includegraphics[scale=0.45]{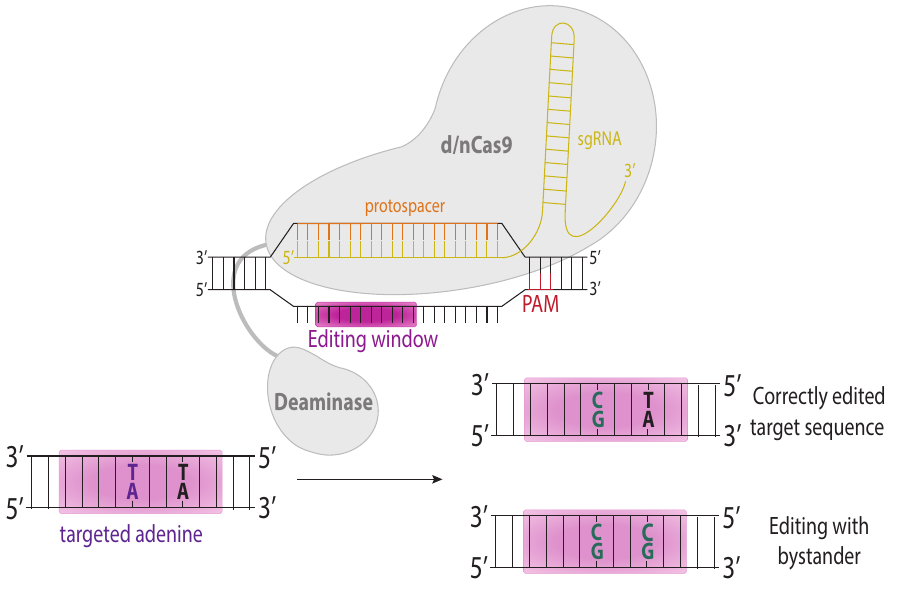}
    \caption{\small{Adenine base editor}} 
    \label{fig:base_editor}
    \centering
    \includegraphics[scale=0.22]{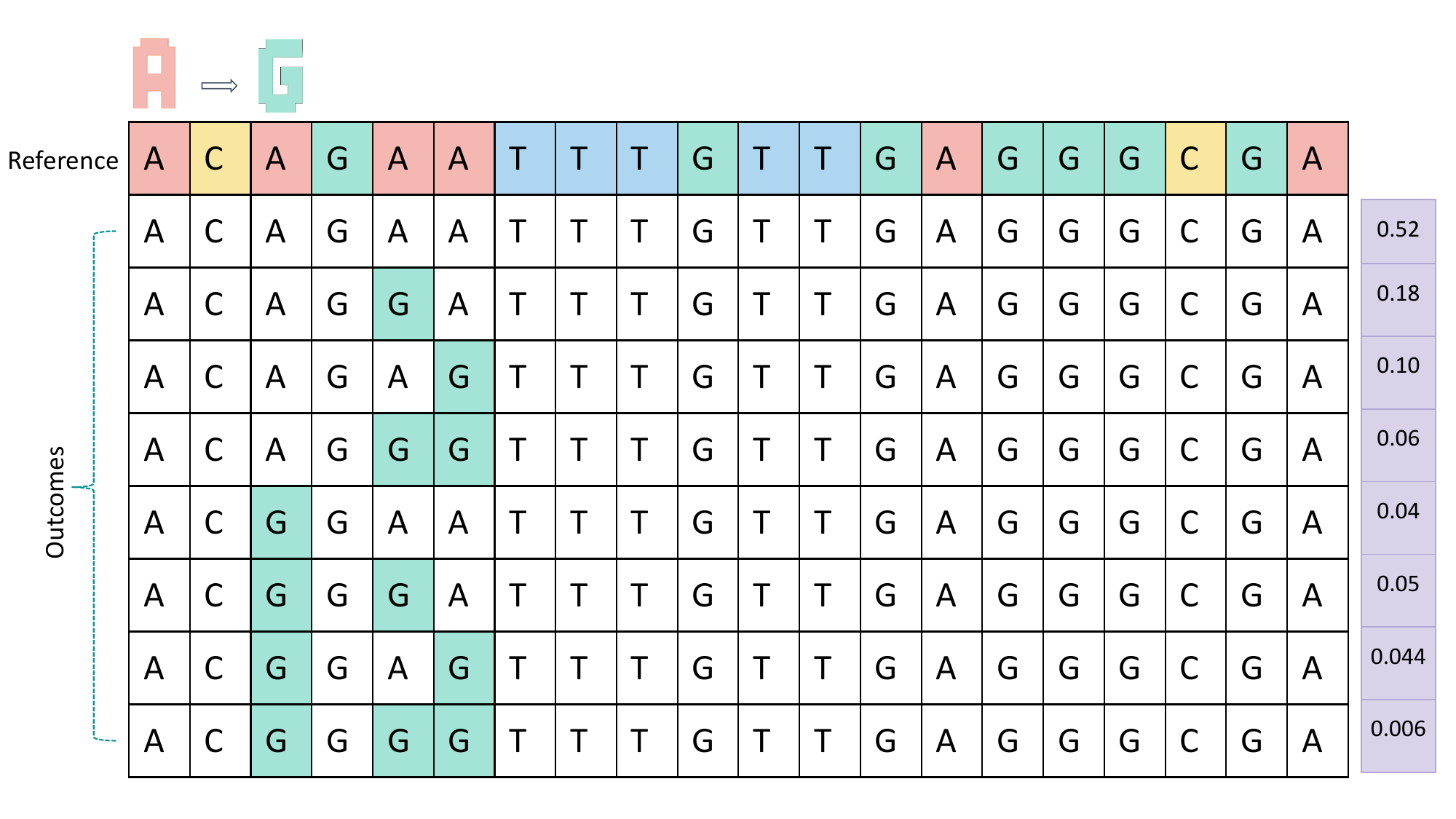}
    \caption{\small{
    An example of a reference sequence of 20 bases (i.e. nucleotides) and associated outcome sequences when applying ABEmax base editor. The first row represents the reference (target) sequence, and the second row is the outcome sequence with no modification (i.e. wild-type) with a probability of occurrence of 0.52. The third row represents a possible outcome sequence where the letter A is changed to G at position 5 with a probability of 0.35. The rest of the rows represent all possible changes of the reference sequence targeting letters A to G with their associated probabilities.}}
    \label{fig:data_demo}
\end{figure}

\paragraph{Base editor and related concepts}
Base editors (BEs) are created by fusing the Cas9 protein with DNA-modifying enzymes. They are directed by a 20-base pair guiding RNA molecule (sgRNA) that acts as a GPS to locate and bind to a matching DNA segment known as the \emph{protospacer}. The effectiveness of BEs largely depends on the composition of this protospacer sequence. BEs, in tandem with the sgRNA, can only bind to the DNA if there's a \emph{protospacer adjacent motif} (PAM) - a sequence consisting of 2-6 nucleotides - present adjacent to the protospacer. This PAM sequence further influences the activity of BEs. There are two primary types of base editors: adenine base editors (ABEs) (presented in figure \ref{fig:base_editor}), which convert adenine (A) to guanine (G), and cytosine base editors (CBEs) that chemically convert cytosine (C) to thymine (T). A detailed description of the base editor is provided in the appendix section \ref{apend:base_editor}.

\subsection{ Data representation} 
Assume we have a target (reference) DNA sequence denoted as $\bx_{\text{ref}} = [x_1, x_2,\dots, x_T]$ where $x_i\in \{A, C, G, T\}$, and a  set of DNA sequences $ \mathbf{X}_{\text{out}} = [\bx_{\text{out},1}, \bx_{\text{out},2}, \dots, \bx_{\text{out}, M}]\in \mR^{M\times T}$ representing corresponding outcomes when a specific base editor is applied to the reference sequence $\bx_{\text{ref}}$. The associated probabilities for these outcomes are given by $\by=[y_1, y_2, \dots, y_{M}]$ where $y_i= P(\bx_{\text{out}, i}|\bx_{\text{ref}}) \in [0,1], \: \textit{for} \:\:i = 1, 2, \dots, M$,  indicating the likelihood of obtaining outcome $\bx_{\text{out},i}$ through editing of $\bx_{\text{ref}}$. Here, $T$ is the length of the reference sequence, and $M$ represents the total number of possible outcomes for a given reference sequence. The count of outcomes can vary depending on the reference sequence. An example of a reference sequence and associated outcome sequences is represented in Figure \ref{fig:data_demo}.

In this paper, we use bold uppercase letters for matrices ($\mathbf X$), bold lowercase letters for vectors or sequences ($\bx$), and regular non-bold letters for scalars or tokens. We use $P$ for probability distributions and non-bold uppercase letters ($X$) for random variables. 
To represent the reference sequence, 
we consider protospacer, PAM, and overhangs. Here, "overhangs" refer to adjacent nucleotides on both sides of the protospacer.. 
To declutter the notation, we will mainly use $\bx_{\text{ref}}$ to denote the reference sequence which could refer to one of these representations: (a) protospacer, (b) protospacer + PAM, or a (c) left overhangs + protospacer + PAM + right overhangs where + is the concatenation operator. Respectively, the outcome sequences are the DNA sequences with the same length as the reference sequence and with a modification of the target bases at the protospacer. The outcome sequence identical to the reference sequence (no edits) is referred as the wild-type.

The training dataset comprises $N$ pairs, each containing a reference sequence, its associated outcomes, and the corresponding probabilities, denoted as $D = \{\bx_{\text{ref}}^i, \mathbf X_{\text{out}}^i, \by^i\}_{i=1}^N$. For simplicity, when referring to a specific reference sequence and its outputs, we omit the instance-level indexing and use only $\bx_{\text{ref}}$.

\subsection{Problem formulation}\label{seq: model_formulation}
Our objective is to predict the likelihood of potential outcomes resulting from a specific base editor applied to a reference sequence.
One approach would be formulating it as a generative model where we directly model the condition distribution $P(X_{\text{out}}|\bx_{\text{ref}})$ that we can both sample different outcomes for a given reference sequence and calculate the probability of each outcome. However, unlike typical generative models that must learn to generate entire output sequences, our scenario benefits from already knowing a portion of the output sequences. Due to the base editor's specific targeting of A-to-G or C-to-T transformations, a substantial portion of the output sequence remains consistent with the reference sequence, with only a few positions undergoing alteration. 

In the inference phase, for a given reference sequence, we can efficiently generate all possible outcomes by considering only the edit combination of target bases (A/G) within the protospacer. By traversing through a range of possible edits, we cover the entire landscape of potential outcome sequences. Therefore, we only need to learn the distribution $P(X_{\text{out}}|\bx_{\text{ref}})$ such that we can evaluate the probability of a specific outcome for a given reference sequence $P(X_{\text{out}} = \bx_{\text{out}, i}|\bx_{\text{ref}})$.

\begin{figure}[tb]
    \centering
    \includegraphics[width=9cm, height=10cm]{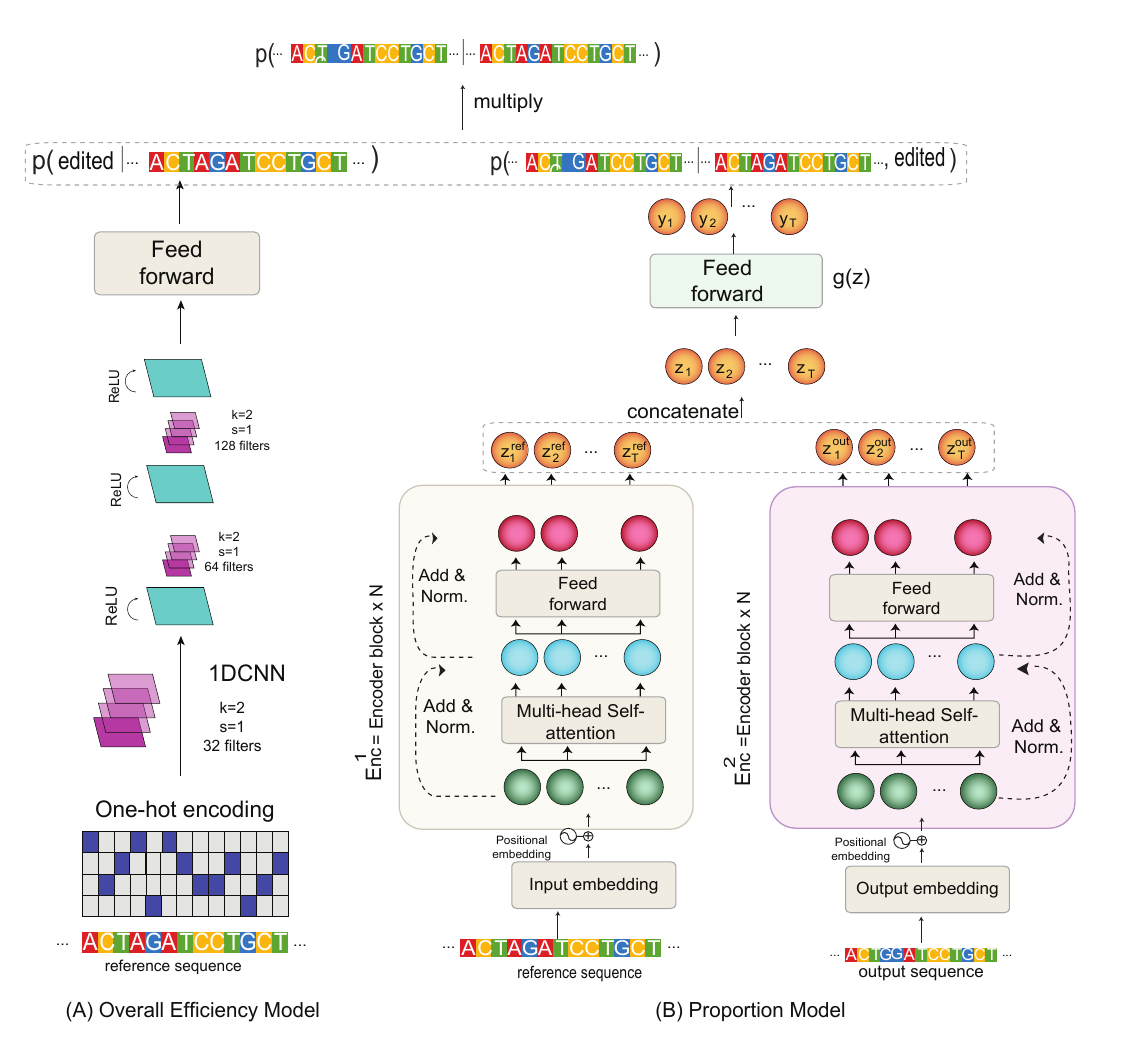}
    
    \caption{Two-stage Model overview}
    \label{fig:Two-step-Model}
\end{figure}

\paragraph{One-stage Model}
In this setup, we tackle the problem by learning a function $f(\bx_{\text{ref}}, \bx_{\text{out}, i})\rightarrow \hat{y}_i$ where $i = 1, \dots, M$, and $\sum_{i=1}^M\hat{y}_i=1$, that takes as input the reference sequence and one of its corresponding outcome and learns to approximate the probability of obtaining that specific outcome. Notably, this function $f$ characterizes a categorical distribution $P(X_{\text{out}} = \bx_{\text{out}, i}|\bx_{\text{ref}})\sim Cat(M, \hat{\by})$, where $\hat{\by}$ is the vector containing probabilities for M outcomes. 
To learn the function $f$, we propose to use attention-based encoder blocks to learn the encoding of both the reference sequence and output sequence. Subsequently, we apply a prediction model on the learned encoded representation to output the probability of obtaining the outcome. The network architecture to learn $f$ is reported in figure \ref{fig:Two-step-Model} (B: proportion model). However, there is a relatively higher probability often associated with the wild-type outcome 
($\bx_{\text{out}, i}=\bx_{\text{ref}}$), while the probabilities linked to the edited outcome sequences are often very small. This situation presents a challenge when directly modeling $P(X_{\text{out}}|x_{\text{ref}})$— as the model might easily learn the wild-type probability but struggle with outcomes that have extremely low probabilities. 

\subsection{Two-stage model}
To address this, we propose a two-stage model where we break down $P(X_{\text{out}}|\bx_{\text{ref}})$ as the product of two probabilities:

\begin{multline}
P(\bx_{\text{out},i}|\bx_{\text{ref}}) =
\begin{cases}
    P(\bx_{\text{out}, i}|\bx_{\text{ref}}, \text{edited})P(\text{edited}|\bx_{\text{ref}}), \\ \:\:\:\:\:\:\:\:\:\:\:\:\:\:\:\:\:\:\:\:\:\:\:\:\:\:\:\:\:\:\:\:\:\:\:\:\:\text{if } \bx_{\text{out}, i} \neq \bx_{\text{ref}} \\
    1 - P(\text{edited}|\bx_{\text{ref}}), \text{if } \bx_{\text{out}, i} = \bx_{\text{ref}}
\end{cases}
\label{model:full}
\end{multline}

For a given reference sequence, we first predict the probability of overall efficiency which is defined in Eq. \ref{eq:efficiency}. It provides the probability of the target sequence being edited, $P(edited|\bx_{\text{ref}})$, which in turn gives the probability of the wild-type. Next, we predict the probability of all possible edited outcomes, $P(\bx_{\text{out}, i}|\bx_{\text{ref}}, edited)$. We refer to the first as \textit{overall efficiency} and the second as \textit{proportion}
\begin{equation}
\resizebox{0.9\linewidth}{!}{
 $P(edited|\bx_{\text{ref}})= \frac{\textit{Sum of the read count of all edited reads for the target}}{\textit{Total read count of the target sequence}}$
 }
 \label{eq:efficiency}
\end{equation}

We estimate the overall efficiency of the given reference sequence using $f_{\mathbf{\theta}_1}(\bx_{\text{ref}})$ (Eq. \ref{eq:model_overalleff}), denoted by the overall efficiency model, and subsequently, we predict the conditional probabilities of all non wild-type outcomes using $f_{\mathbf{\theta}_2}(\bx_{\text{ref}},\bx_{\text{out}, i})$ (Eq. \ref{eq:model_bystander}) which we denote by the proportion model.

\begin{equation}
 f_{\mathbf{\theta}_1}(\bx_{\text{ref}}) = P(edited|\bx_{\text{ref}})
 \label{eq:model_overalleff}
 \end{equation}
 $ \text{ where}\:\:P(\textit{wild-type}|\bx_{\text{ref}})  = 1-P(edited|\bx_{\text{ref}}) $
 \begin{equation}
 f_{\mathbf{\theta}_2}(\bx_{\text{ref}},\bx_{\text{out}, i}) = P(\bx_{out, i}|\bx_{\text{ref}}, \textit{edited}), \:\: 
  \label{eq:model_bystander}
\end{equation}
$\text{where}\:\:  \bx_{\text{out}, i}\neq \bx_{\text{ref}}$

Once $f_{\mathbf{\theta}_1}$ and $f_{\mathbf{\theta}_2}$ are learned,  we can calculate $P(X=\bx_{\text{out}, i}|\bx_{\text{ref}})$ where $i = 1, \dots M$ for all outcome sequences, including wild-type and edited sequences using Eq \ref{model:full}.

\subsubsection{Overall efficiency model}
We formulate the overall efficiency model as a probabilistic classification task where $f_{\mathbf{\theta}_1}$ parameterizes a binomial distribution $P(C|\bx_{\text{ref}})$ of a random variable $C\in \{\textit{edited}, \textit{not edited}\}$ with the aim to learn to output the $P(C=edited|\bx_{\text{ref}})$ for a given reference sequence. 
To learn $f_{\mathbf{\theta}_1}$, we first computed the overall editing efficiency for each reference sequence by summing all probabilities attributed to the non wild-type outcomes as given in Eq \ref{eq:efficiency}, or equivalently, $1-P(\textit{wild-type} |\bx_{ref})$. 
Then we use multiple 1D-Convolutional layers \citep{lecun1995convolutional} on the one-hot-encoded representation of $\bx_{ref}$ to learn discriminative feature embedding that is passed to the multi-layer perceptron (MLP) layer to approximate the distribution $P(C|\bx_{\text{ref}})$. 
The model architecture is presented in Figure \ref{fig:Two-step-Model} (A).
We trained $f_{\mathbf{\theta}_1}$ using KL-divergence loss that is applied on the true distribution $P(C|\bx_{\text{ref}})$  and learned distribution $\hat{P}_{\theta_1}(C|\bx_{\text{ref}})$ for each reference sequence.
\begin{align}
\resizebox{0.9\linewidth}{!}{
 $\mL_{\textit{efficiency}}(\theta_1, D) = \sum_{i=1}^N D_{kl} (P(C|\bx_{\text{ref}}^i)\|\hat{P}_{\theta_1}(C|\bx_{\text{ref}}^i))$}
\end{align}

\subsubsection{Proportion model}
This model is designed  to approximate the conditional distribution $P(X_{\text{out}}|\bx_{\text{ref}}, \textit{edited})$. To achieve this, we first remove the wild-type from each reference sequence's corresponding output $X_{\text{out}}$. Then, we normalize the probabilities of the remaining outcomes 
to ensure a valid distribution effectively converting $P(X_{\text{out}}|\bx_{\text{ref}})$ into the distribution $P(X_{\text{out}}|\bx_{\text{ref}}, \textit{edited})$. 
 The proportion model $f_{\mathbf{\theta}_2}$ is designed to learn  the parameters governing the  distribution $P(X_{\text{out}}|\bx_{\text{ref}}, \textit{edited})$. Similar to the one-stage model, $f_{\mathbf{\theta}_2}$ is provided with both the reference sequence $\bx_\text{ref}$ and its associated outcome sequence $\bx_{\text{out}, i}$. The model is then trained to estimate the likelihood $P(\bx_{\text{out}, i}\:|\:\bx_{\text{ref}}, \textit{edited})$, representing the probability of reference sequence being edited, and result in the outcome sequence $\bx_{\text{out}, i}$.

As illustrated in Figure \ref{fig:Two-step-Model} (B), $f_{\mathbf{\theta}_2}$ uses attention-based model comprised of two encoder networks, $\text{Enc}^{1}(\bx_{\text{ref}})$, $\text{Enc}^{2}(\bx_{\text{out}})$, and one output network $g$. The design of the encoder networks adapts the transformer encoder blocks architecture \citep{vaswani2017attention}, characterized by multiple layers of multi-head self-attention modules. 
The two encoder networks process the reference sequence and one of its corresponding output sequence $\bx_{\text{out}, i}$, leading to the extraction of their respective latent representations, namely $\mathbf Z_{\text{ref}}\in \mR^{T\times d}$ and $\mathbf Z_{\text{out}}\in \mR^{T\times d}$. Both vectors are then concatenated to form a unified learned representation $\mathbf Z\in\mathbb{R}^{T\times 2d}$. 
Subsequently, the output network $g$ embeds this unified representation $\mathbf Z$ to compute the probability of obtaining the output sequence given the reference sequence,  $P(\bx_{\text{out}, i}\:|\:\bx_{\text{ref}}, \textit{edited})$.

 Precisely,  the output network $g(\mathbf Z)$ takes as input the final representation $\mathbf Z\in\mathbb{R}^{T\times 2d}$  and performs an affine transformation followed by softmax operation to compute the probability of conversion  of every target base (i.e. base A or C depending on the chosen base editor) as it is shown below:
\begin{equation}
\hat{y}_{it}  = \sigma(\mathbf W\mathbf z_{it} + \mathbf b_t)
\end{equation}
where $\mathbf W\in\mR^{2\times 2d}, \mathbf b_t\in\mR^2$ and   $\sigma$ is softmax function. $\hat{y}_{it}$ represents the probability of editing occurring at the $t$-th position in the $i$-th outcome sequence. The un-normalized probability for the whole $i$-th output sequence $\bx_{\text{out}, i}$ given its reference sequence is computed by $\hat{y}_i=\prod_{t=1}^T\hat{y}_{i,t}$, which is then normalized across all the outcomes to make it valid probability distribution (Eq. \ref{eq:normalizing_prop}). Therefore, the approximated probability for obtaining $i$-th edited (non-wild type) outcome sequence is given by: 
 \begin{equation}
     \hat{P}(\bx_{\text{out}, i}\:|\:\bx_{\text{ref}}, \textit{edited}) =\frac{\hat{y}_i}{\sum_{i=1}^M\hat{y}_i}
    \label{eq:normalizing_prop}
 \end{equation}

\paragraph{Objective Function}
We used the Kullback–Leibler (KL) divergence on the model’s estimated distribution over all outcome sequences for a given reference sequence $\bx_{\text{ref}}^i$ and the actual distribution:
\begin{flalign}
    &D_{\text{KL}}^i(P(X_{\text{out}}|\bx_{\text{ref}}^i, \textit{edited})||\hat{P}_{\theta_2}(X_{\text{out}}|\bx_{\text{ref}}^i, \textit{edited}))  \\
    &= \sum_{j=1}^{M_i} P(\bx_{\text{out},j}|\bx_{\text{ref}}^i, \textit{edited})\log \frac{P(\bx_{\text{out},j}|\bx_{\text{ref}}^i, \textit{edited})}{\hat{P}_{\theta_2}(\bx_{\text{out},j}|\bx_{\text{ref}}^i, \textit{edited})} \nonumber
\end{flalign}
Lastly, the objective function for the whole training set is defined by the average loss across all the reference sequences as follows:
\begin{align}
    &\mL_{\text{proportion}}(\mathbf{\theta_2}; D) =\\
    &\sum_{i = 1}^N D_{KL}^i (P(X_{\text{out}}\:|\:\bx_{\text{ref}}^i, \textit{edited})||\hat{P}_{\theta_2}(X_{\text{out}}\:|\:\bx_{\text{ref}}^i, \textit{edited}) \nonumber
\end{align}

The final objective is composed of both the overall efficiency model loss and the proportion model loss with a weight regularization term (i.e. $l_2$-norm regularization) applied to the model parameters represented by $\mathbf{\theta} = \{\mathbf{\theta_1}, \mathbf{\theta_2}\}$ (Eq. \ref{eq:composite_loss_func})
\begin{equation}
  \mL_{\textit{proportion}}(\mathbf{\theta_1}; D) + \mL_{\textit{efficiency}}(\mathbf{\theta_2}, D)+ \frac{\lambda}{2}\|\mathbf \theta\|_2^2
  \label{eq:composite_loss_func}
\end{equation}

\subsection{Multi-task learning with multiple base editors}\label{sec:Multi_task_learning}
\begin{figure}[H]
    \centering
    \includegraphics[width=8cm, height=5cm]{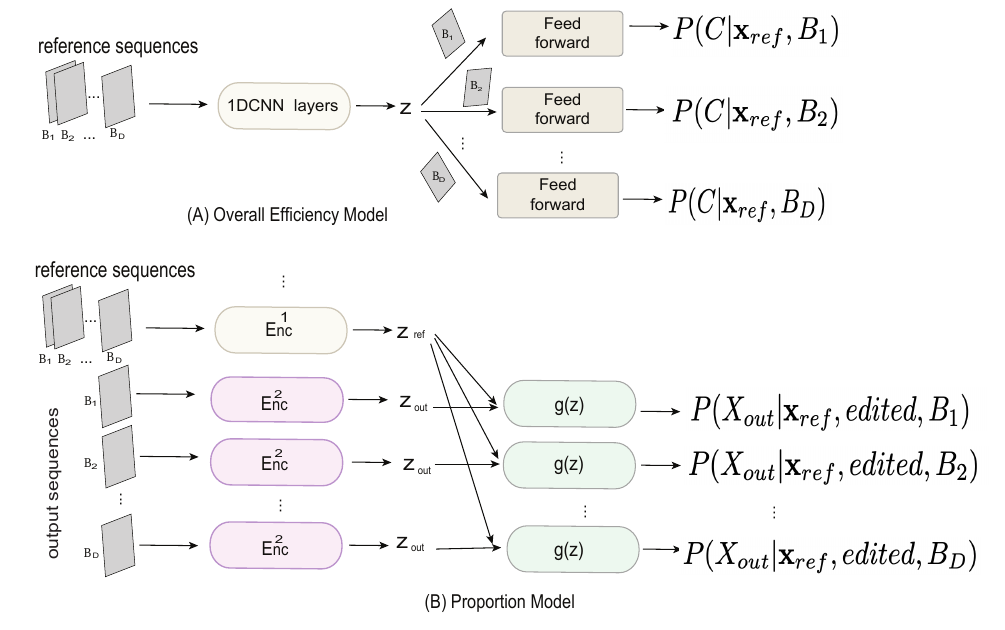}
    \caption{Multi-task learning model overview}
    \label{fig:multi_task}
\end{figure}
\paragraph{Multi-task learning} There is a diverse set of base editors, each distinguished by its unique design attributes. These distinctions, including variations in binding affinities, editing window sizes, and deaminase activities, result in differing editing efficiencies even when targeting the same sequence. The variations in editing efficiency across different editors emphasize the complexity of the base editing landscape. 
Conventional approaches have often proposed training separate models for each editor. However, this approach not only demands additional effort but also fails to leverage the shared structural similarities among editors. 
To leverage common patterns and relationships present across various libraries derived from different base editors, and optimize predictive capability while also reducing computational time, we propose a more efficient solution based on multi-task learning. Instead of training separate models for each editor, we train a single model capable of predicting the efficiency of all editors when applied to the same reference sequence. 

 Given a total number of $D$ editors where each editor has its own dataset $B_i$ (denoted by screening libraries), we developed a multi-task learning model that uses shared encoding layers to extract a common representation across all the libraries as well as individual branches that fine-tune the model specifically for each library, ensuring a better fit to their respective data. 
 This approach implicitly models $P(X_{\text{out}}\:|\:\bx_{\text{ref}}, B_i)$ where $B_i$ represents the base editor type applied on the reference sequence.
To implement one universal model across all datasets,  we extend our proposed two-stage model architecture (illustrated in Figure \ref{fig:Two-step-Model}) for multi-task learning, as depicted in Figure \ref{fig:multi_task}. 

Specifically, we modify the overall efficiency model by initially employing two convolutional layers as a shared building block across all datasets/editors, enabling the learning of a common representation for the reference sequence.  Then a set of output blcoks is used to represent editor-specific transformation. Each editor type has its own output block consisting of two layers of convolutional network followed by MLP layers to predict the probability $P(edited|\bx_{\text{ref}})$ for each editor/dataset accordingly.

We adapt the proportion model by using a common encoder network across all editors/datasets to establish a unified representation $\mathbf Z_{\text{ref}}$ for the reference sequence while using separate encoders and output blocks for each distinct editor. To counterbalance any bias towards larger datasets, we implemented a data loader that uniformly samples the same number of data samples in each mini-batch throughout the training phase.
\section{Experiments}
\subsection{Dataset and Experiment setup}
\paragraph{Dataset} 

To comprehensively assess base editors' efficiency across thousands of genomic sequences, we conducted high-throughput screening, resulting in the creation of six distinct datasets. Each dataset corresponds to the application of one of the following base editors: SpRY-ABE8e, SpCas9-ABE8e SpG-ABE8e, SpRY-ABEmax, SpCas9-ABEmax, and SpG-ABEmax, as listed in Table \ref{tab:data}. Detailed descriptions of the used editors are provided in Appendix Section \ref{appendix:BE}. Each dataset encompasses approximately 11,000 reference sequences and their corresponding output sequences. In each dataset, we leveraged 193 distinct PAM sites, each comprising four nucleotide bases. 

\begin{table}[tb]
    \centering
    \resizebox{\linewidth}{!}{%
    \begin{tabular}{llllll}
    \toprule
         Editor & \#ins&\#refseq& \#outcome&mean 
         &std  \\
         \hline
         SpRY-ABE8e&110141&11291 &9.7&0.102&0.211\\
        
         SpCas9-ABE8e&43054&11337 &4.6&0.217&0.323\\
         
         SpG-ABE8e&80873&11307&7.1&0.139&0.263\\
        SpRY-ABEmax&70851&11347&6.2&0.159&0.301\\
        SpCas9-ABEmax&39606&11302&3.5&0.285&0.417\\
        SpG-ABEmax&70851&11347&6.2&0.159&0.301\\
\bottomrule
    \end{tabular}}
    \caption{\small{Data statistics: ``\#ins" refers to the number of reference and output sequence pairs, ``\#refseq" denotes the number of distinct reference sequences, ``\#outcome" denotes the average number of outcomes per reference sequence, the mean and std refers to the mean and standard deviation of the probability across all the outcomes.}}
    \label{tab:data}
\end{table}
\paragraph{Experiment setup}
We divided every dataset into training, testing, and validation sets, maintaining a ratio of 80\%, 10\%, and 10\%. This procedure is repeated three times to ensure robust performance reporting. All reported results are based on the average performance over the three runs(indicated by $mean \pm std$).  First, we use the one-stage model to identify the best features to represent reference sequence for predicting the base editing outcomes (i.e. determine reference sequence representation option as explained in section \ref{seq: model_formulation}). Using the selected features (i.e., protospacer + PAM), we proceed to evaluate the performance between the one-stage and two-stage models. Finally, using the two-stage model, we compare the multi-task learning (i.e. a unified model training for all editors) to the single-task learning setup where separate models are trained for the different editors.

Throughout model training, we track the epoch at which the best validation scores are attained. Evaluation of the trained models for each base editor is based on their average performance on the test sets across the three runs. Pearson and Spearman correlation were used as performance measures for all tested models. More details about the network structure, optimization, and hyperparameters are presented in the Appendix Section \ref{sec:network_architecture}. 
\subsection{Experiment results}
\paragraph{Reference sequence representation}
Existing models have explored different factors that could affect the base editor's efficiency, which we categorize into three scenarios: 1) the protospacer, 2) the protospacer along with its PAM, and 3) an extended range including left overhangs, protospacer, PAM, and right overhangs. We investigate all three scenarios with the one-stage model to identify the best features to represent the reference sequence. As shown in Table \ref{tab:reference_seq}, we observe that incorporating PAM information significantly enhances performance, whereas the inclusion of overhangs demonstrates minimal impact. Besides, adding overhangs increases the computational complexity drastically. Consequently, we opt to employ protospacer and PAM information to represent reference sequences in all the subsequent model results presented below.

\begin{table*}
    \centering
     \resizebox{0.8\textwidth}{!}{%
    \begin{tabular}{lllllll}
    \toprule
    & \multicolumn{2}{c}{Protspacer}& \multicolumn{2}{c}{Protospacer  \& PAM}&\multicolumn{2}{c}{Protspacer \& PAM \& Overhangs}\\
      Libraries & Spearman& Pearson&Spearman&Pearson&Spearman& Pearson\\
         \hline
         SpRY-ABEmax    &$0.835 \pm 0.007$&$0.981 \pm 0.001$& $0.854 \pm 0.006$&$0.983 \pm 0.001$&$0.854 \pm 0.003$&$0.983 \pm 0.002$ \\
         SpCas9-ABEmax &$0.786 \pm 0.003$&$0.978 \pm 0.002$&$0.881 \pm 0.001$&$0.989 \pm 0.0005$&$0.891 \pm 0.002$&$0.989 \pm 0.001$\\
         SpG-ABEmax&$0.841 \pm 0.002$&$0.985 \pm 0.0007$&$0.866 \pm 0.004$&$0.989 \pm 0.0003$&$0.878 \pm 0.008$&$0.991 \pm 0.0009$\\
        SpRY-ABE8e &$0.776 \pm 0.019$&$0.965 \pm 0.001$&$0.779 \pm 0.0036$&$0.968 \pm 0.002$&
$0.803 \pm 0.008$&$0.967 \pm 0.0003$\\
        SpCas9-ABE8e&$0.762 \pm 0.007$&$0.883 \pm 0.005$&$0.857 \pm 0.007$&$0.945 \pm 0.0006$&$0.862 \pm 0.003$&$0.945 \pm 0.003$\\
        SpG-ABE8e&$0.803 \pm 0.005$&$0.963 \pm 0.002$&$0.820 \pm 0.005$&$0.974 \pm 0.0009$&$0.819 \pm 0.006$&$0.9771 \pm 0.0008$ \\
    \bottomrule
    \end{tabular}}
    
    \caption{\small{Pearson and Spearman correlation using one-stage Model across the three different reference sequence representations. In our experiment,  we chose 5 neighboring nucleotides for both sides to represent the overhangs. }}
    \label{tab:reference_seq}
\end{table*}


\paragraph{Comparing One-stage with Two-stage Model} 
 As detailed in Section \ref{seq: model_formulation}, our model can be conceptualized as either a one-stage model, directly capturing the distribution across all potential outcomes for a given reference, or as a two-stage model. The latter approach involves initially predicting the probability of an edit occurring in the reference sequence, followed by predicting the probabilities of individual edited outcomes. In this section, we present results for both models to illustrate the advantages of the two-stage approach over the one-stage counterpart. For the one-stage model, we use exactly the same architecture as the proportion model from the two-stage model on the original data without preprocessing where we remove the wild type and renormalize the probability for each reference. 

\begin{table}[tb]
    \centering
     \resizebox{\linewidth}{!}{%
    \begin{tabular}{lllll}
    \toprule
    & \multicolumn{2}{c}{One-stage Model}& \multicolumn{2}{c}{Two-stage Model}\\
      Libraries & Spearman& Pearson&Spearman&Pearson\\
        
         \hline
         SpRY-ABEmax &$0.854\pm 0.006$& $0.983\pm 0.001$& $0.873 \pm 0.001$ & $ 0.986 \pm 0.001$ \\
        
         SpCas9-ABEmax &$0.881\pm 0.006$&$0.989\pm 0.0005$  & $0.879 \pm 0.004$ & $0.991 \pm 0.001$ \\

         SpG-ABEmax &  $0.866 \pm 0.004$ & $0.989 \pm 0.0003$& $0.887\pm 0.003$ &$0.991\pm 0.0006$\\
        SpRY-ABE8e &  $0.779 \pm 0.003$ & $0.968 \pm 0.002$ &$0.862\pm 0.003$ &$0.974\pm 0.001$\\
        SpCas9-ABE8e &$0.857\pm 0.007$&$0.945\pm 0.0006$& $0.856 \pm 0.003$ & $0.937 \pm 0.002$  \\
        SpG-ABE8e & $0.820\pm 0.005$&$0.974\pm 0.0009$ &$0.865 \pm 0.004$ & $0.978 \pm 0.0008$ \\
    \bottomrule
    \end{tabular}}
    \caption{\small{Prediction performance all outcomes (i.e. including wild-type sequences).}}
    \label{tab:two-step-model_overlall}

    \centering
     \resizebox{\linewidth}{!}{%
    \begin{tabular}{lllll}
    \toprule
    & \multicolumn{2}{c}{One-stage Model}& \multicolumn{2}{c}{Two-stage Model}\\
      Libraries & Spearman& Pearson&Spearman&Pearson\\
        
         \hline
         SpRY-ABEmax &$0.745\pm 0.015$& $0.711\pm 0.011$& $0.799\pm 0.007$ & $0.782 \pm 0.012$ \\
        
         SpCas9-ABEmax &$0.82\pm 0.0003$&$0.851\pm 0.014$  & $0.838 \pm 0.009$ & $0.890 \pm 0.030$ \\

         SpG-ABEmax &  $0.807 \pm 0.003$ & $0.752 \pm 0.014$& $0.845\pm 0.011$ &$0.822\pm 0.014$\\
        SpRY-ABE8e &  $0.393 \pm 0.021$ & $0.508 \pm 0.025$ &$0.547\pm 0.056$ &$0.669\pm 0.051$\\
        SpCas9-ABE8e &$0.855\pm 0.007$&$0.840\pm 0.003$& $0.866 \pm 0.0021$ & $0.858 \pm 0.021$  \\
        SpG-ABE8e & $0.712\pm 0.002$&$0.732\pm 0.004$ &$0.774 \pm 0.005$ & $0.810 \pm 0.009$ \\
    \bottomrule
    \end{tabular}}
    \caption{\small{Prediction performance on the wild-type outcomes}}
    \label{tab:two-step-model_wild}
    \centering
     \resizebox{\linewidth}{!}{%
    \begin{tabular}{lllll}
    \toprule
    & \multicolumn{2}{c}{One-stage Model}& \multicolumn{2}{c}{Two-stage Model}\\
      Libraries & Spearman& Pearson&Spearman&Pearson\\
        
         \hline
         SpRY-ABEmax &$0.740\pm 0.007$& $0.778\pm 0.012$& $0.798\pm 0.003$ & $0.818 \pm 0.006$ \\
        
         SpCas9-ABEmax &$0.683\pm 0.0003$&$0.748\pm 0.022$  & $0.728 \pm 0.006$ & $0.795 \pm 0.006$ \\

         SpG-ABEmax &  $0.729 \pm 0.0043$ & $0.744 \pm 0.004$& $0.778\pm 0.005$ &$0.810\pm 0.004$\\
        SpRY-ABE8e &  $0.707 \pm 0.010$ & $0.816 \pm 0.006$ &$0.809\pm 0.004$ &$0.849\pm 0.003$\\
        SpCas9-ABE8e &$0.684\pm 0.007$&$0.729\pm 0.008$& $0.714 \pm 0.014$ & $0.753 \pm 0.007$  \\
        SpG-ABE8e & $0.719\pm 0.004$&$0.787\pm 0.005$ &$0.789 \pm 0.004$ & $0.826 \pm 0.003$ \\
    \bottomrule
    \end{tabular}}
    \caption{\small{Prediction performance on the non wild-type outcomes (i.e. edited outcome sequences)}}
    \label{tab:two-step-model_nonwild}
\end{table}
Table \ref{tab:two-step-model_overlall} shows the two-stage model has slightly superior results (Spearman correlation) over the one-stage model. This improvement can be attributed to the model's two-step prediction approach, 
which first predicts the wild-type alone and subsequently refines predictions for various edited outcomes. To better understand the difference between the two models' ability to predict the wild-type and edited outcome sequences, we rigorously evaluated each model's performance separately on both types of outcome. The two-stage model outperforms the one-stage model in most of the datasets when considering both wild type and edited outcomes as presented in Table \ref{tab:two-step-model_wild} and \ref{tab:two-step-model_nonwild}.
\paragraph{Multi-task learning}
Given this conclusion, we proceed with the multi-task learning with the two-stage mode (see  Figure \ref{fig:multi_task}). We compared the performance of multi-task learning across all the datasets/editors with a single-task setup where we trained one model per dataset/editor.  Table \ref{tab:muti-task} reports similar performance for both models. Although there wasn't a substantial performance difference, adopting a unified multi-task model offers advantages such as reduced run-time (for training and inference) and smaller model size (fewer parameters) while maintaining consistent performance across all datasets. Moreover, with a unified model, we can simultaneously predict the editing outcomes of all six editors at once for a given target sequence.

\begin{table}[tb]
    \centering
     \resizebox{\linewidth}{!}{%
    \begin{tabular}{lllll}
    \toprule
    & \multicolumn{2}{c}{Single task learning}& \multicolumn{2}{c}{Multi task learning}\\
      Libraries & Spearman& Pearson&Spearman&Pearson\\
        
         \hline
         SpRY-ABEmax & $0.877 \pm 0.001$ & $0.986 \pm 0.001$ & $0.872 \pm 0.002$ & $0.986 \pm 0.0002$ \\
        
         SpCas9-ABEmax & $0.879 \pm 0.004$ & $0.989 \pm 0.001$ & $0.864 \pm 0.0019$ & $0.992 \pm 0.0001$ \\

         SpG-ABEmax & $0.882 \pm 0.001$ & $0.991 \pm 0.0006$ & $0.889 \pm 0.0016$ & $0.992 \pm 0.0004$ \\
        SpRY-ABE8e & $0.861 \pm 0.0029$ & $0.974 \pm 0.001$ & $0.863 \pm 0.0011$ & $0.975 \pm 0.001$ \\
        SpCas9-ABE8e & $0.856 \pm 0.008$ & $0.938 \pm 0.0005$ & $0.852 \pm 0.002$ & $0.937 \pm 0.003$ \\
        SpG-ABE8e & $0.865 \pm 0.004$ & $0.980 \pm 0.0008$ & $0.871 \pm 0.003$ & $0.979 \pm 0.001$ \\
    \bottomrule
    \end{tabular}}
    \caption{\small{Performance comparison between the multi-task and single task learning models}}
    \label{tab:muti-task}
\end{table}

\paragraph{Comparing to baselines in the literature}
We also compared our model with  BE-DICT \citep{marquart2021predicting} which is one of the relevant existing models that tackle base editing outcome prediction. 
BE-DICT is a sequence-to-sequence model where the decoding happens in an auto-regressive manner, it is computationally heavy compared to our proposed method. Moreover, it is trained as single-task model (i.e. one model for each editor) and use only the protospacer to represent the target sequence. We extended and retrained BE-DICT on two of the datasets (randomly chosen) and compared the prediction results with ours. For a fair comparison, we first used our one-stage model trained in the single task setting (one model per dataset) using only the protospacer. The results of this experiment reveal the advantages of the architectural changes, particularly in adopting an encoder-encoder architecture over the traditional sequence-to-sequence (encoder-decoder) model. 
Then we extended their model to use both the protospacer and PAM as the reference sequence and compared it with our proposed multi-task model (trained on eight data sets using protospacer and PAM information as target sequence).


\begin{table}[tb]
\centering
     \resizebox{\linewidth}{!}{%
    \begin{tabular}{llllll}
    \toprule
    && \multicolumn{2}{c}{ BE-DICT}& \multicolumn{2}{c}{Ours}\\
     reference sequence & Libraries & Spearman& Pearson&Spearman&Pearson\\  
         \hline
    \multirow{2}{*}{prrotospacer}& SpRY-ABEmax & 0.801  & 0.943&0.835 &0.981\\
                                  &SpRY-ABE8e & 0.746&0.861 &0.776&0.965 \\
    \multirow{2}{*}{prrotospacer \& PAM}&SpRY-ABEmax & 0.804 &0.951 &0.870 &0.987 \\
    &SpRY-ABE8e &0.762&0.850&0.860&0.975\\
    \bottomrule
    \end{tabular}}
    \caption{\small{Performance comparison with the baselines}}
    \label{tab:baselines}
\end{table}

Results in Table \ref{tab:baselines} show that our model consistently outperforms BE-DICT. Furthermore, considering computational efficiency during model training (on SpRY-ABE8e data using the protospacer and PAM reference sequence representation), BE-DICT takes in the order of minute per epoch (wall time), while our single-task model accomplishes the same task in the order of seconds (~15 seconds). Notably, the multi-task learning model trained jointly on all six datasets takes ~21 seconds per epoch.

This highlights the benefits of avoiding the complex sequence-to-sequence architecture in favor of a streamlined encoder-encoder structure. This choice not only improves the computational efficiency but also leads to model predictive performance improvements. Moreover, the introduction of a two-stage model and a multi-task framework amplifies these performance gains even further. We present additional results for comparisons with other baselines in Table \ref{tab:baseline_2} in the appendix. 

To assess our model's performance against other state-of-the-art models, we conducted evaluations using the test sets provided by these models. Table \ref{tab:baseline_2} displays our findings, which include three most recent models: BE-HIVE \citep{arbab2020determinants}, DeepABE \citep{song2020sequence}, and BEDICT \citep{marquart2021predicting}, along with their respective test sets labeled as A. et al., S. et al., and M. et al.

\begin{table}[tb]
    \centering
    \resizebox{\linewidth}{!}{%
    \begin{tabular}{lllllll}
    \toprule
    &\multicolumn{3}{c}{ All Outocmes}& \multicolumn{3}{c}{Non wild-types}\\
         Datasets& A.et all&S. et al &M. et al& A.et all&S. et al &M. et al  \\
         \hline
     BEDICT & 0.96& 0.94&0.86&0.81& 0.90&0.82\\
     DeepABE&0.86&0.93&0.8&0.86&0.96&0.84\\
     BE-HIVE&0.71&0.88&0.74&0.92&0.93&0.81\\
     Our model & 0.972&0.974&0.972& 0.939&0.945&0.953\\
     \hline
    \end{tabular}}
    \caption{\small{Model performance on the test set from the different published studies. Columns represent test sets, rows represent models used}}
    \label{tab:baseline_2}

\end{table}
The idea is to take the published trained model and evaluate their performance on those various test sets. For the three baseline models, we refer to the results reported in the BEDICT paper. As for our model, to ensure fairness in comparison, we used our single-stage model trained on SpG-ABEmax libraries 
since most baselines, except DeepABE, do not incorporate the PAM as input. The results correspond to two scenarios: 1) considering all possible outcomes, and 2) only considering non-wild type outcomes. The results for the non-wild type outcomes correspond to the model prediction where we only consider non-wild outcomes.  In the case of non-wild-type outcome prediction, we mention that other models were trained exclusively on non-wild outcomes, with outcomes per sequence being renormalized. Our one-stage model, however, was trained on data encompassing all outcomes, so we report non-wild-type results with outcomes renormalized for a fair comparison.

\section{Conclusion}
Our work provides a detailed assessment of the modeling approaches for base editor outcome prediction. Through the development of a unified model, we transcend the limitations of single-editor models and pave the way for more versatile and comprehensive tools. By combining self-attention mechanisms and multi-task learning, we capture the nuances of editing outcomes across various editors, enhancing the accuracy and applicability of our predictions.

As the first machine learning-focused paper in the domain of base editor outcome prediction, our work represents a stepping stone toward a more systematic and informed modeling approach to genome editing. We explored the different modeling decisions from one-stage to two-stage models, and from single-task to multi-task learning. We evaluated the different sequence representations and benchmarked our best model with one of the main models developed for base editing outcome prediction.
We believe that further work studying systematically the different modeling decisions for genome editing will help guide researchers toward more promising editing strategies that in turn will bring advancements in gene therapy and disease modeling.

For the future, given the current absence of standardized and systematic benchmark datasets in the field, we aim to bridge this gap by creating standard benchmark datasets, establishing baseline models, and proposing better performance metrics. This initiative will provide the machine-learning community with a solid foundation for testing a wide array of innovative ideas and approaches.
\section*{Acknowledgements}
We thank G. Schwank, K. Marquart, L. Kissling and S. Janjuha for input on the CRISPR-Cas and base editing technology and for data sharing and preprocessing. This work was supported by the URPP ‘Human Reproduction Reloaded’ and ‘University Research Priority Programs’.
\clearpage
\bibliography{example_paper}
\bibliographystyle{icml2022}
\clearpage
\section{Appendix}
\subsection{CRISPR related terminology}\label{crispr}
The CRISPR-Cas9 system is a revolutionary gene-editing technology that allows scientists to precisely modify DNA within living organisms. Initially, it was described as an adaptive immune system of bacteria and archea to eliminate invading foreign DNA and/or RNA. The system uses unique sequences of RNA called single guide RNA (sgRNA) that are recognized and bound by Cas9, an enzyme with a nuclease domain. The Cas9 protein carries out the initial steps of recognition and binding by scanning genomic DNA to locate a particular sequence called a PAM (protospacer adjacent motif). Upon PAM recognition, the part of the sgRNA (termed spacer) complementary to the target DNA (termed protospacer) opens the DNA double helix and binds to the target site. This leads to a conformation change within the Cas9, bringing its nuclease domain in close proximity to the target DNA and thus initiating DNA double-strand cleavage.
After introducing a DNA double-strand break, the cell's repair mechanism is triggered, which can lead to various outcomes. Researchers can exploit this repair process to either introduce specific changes in the DNA sequence by providing a modified DNA template or to disrupt a target gene due to the imperfect repair of the DNA by the cells, thus introducing insertions or deletions, which can lead to a frame shift mutations.

Of note, Streptococcus pyogenes Cas9 (SpCas9) exclusively operates on ``NGG'' (``N'', any base) PAM sequence. Recent efforts in protein design have resulted in laboratory-generated SpCas9 variants, such as SpG or SpRY, which are able to recognize different PAM motifs. Therefore, the editors used in our high-throughput screening are configured to operate with PAM sequences comprising four nucleotides.


\subsubsection{Base Editor (BE)}\label{apend:base_editor}
Base editing \citep{komor2016programmable, gaudelli2017programmable, rees2018base} is a second-generation genome editing approach that uses components from CRISPR systems together with other enzymes to directly install point mutations into genomic DNA without making double-stranded DNA breaks (DSBs). BEs comprise a Cas protein with a catalytically impaired nuclease domain fused to a nucleobase deaminase. Similar to the Cas9 nuclease, BEs are directed to the target DNA by the programmable sgRNA and are able to directly convert substrate bases in a specific 'editing window' within the protospacer.




\subsection{Description of Each Base Editor Used in the Experiment}\label{appendix:BE}
There are two main factors that are crucial for the efficiency of the base editor: binding affinities and deaminase activity. 
Binding affinities dictate how effectively an editor is able to identify and interact with specific target sites on the reference sequence. Editors with higher binding affinities tend to exhibit increased accuracy in achieving the desired base modification. Deaminase activity is defined by the type of deaminase used. Additionally, the deaminase also defines editing window size, which refers to the span of nucleotides that an editor can modify around its target site. Editors with larger editing windows can potentially influence a broader range of nucleotides, resulting in increased flexibility in terms of target selection and outcomes.

Here we describe six editors that we used in our screening experiment. We used three different Cas9 orthologs, which show different binding affinities to different PAMs: SpCas9, SpG, and SpRY. SpCas9 recognizes only a few PAMs but shows a very high affinity for those PAMs. SpRY shows the broadest PAM recognition, however a lower affinity for all of them. SpG recognizes more PAM than SpCas9, however less than SpRY and shows lower affinity than SpCas9 but higher as SpRY. For deaminases we used two evolved adenine deaminases named ABEmax and ABE8e, where ABE8e has higher deaminase activity and a bigger editing window size. The combination of the three Cas9 orthologs and the two adenine deaminases leads to a total of six adenine base editors.



\subsection{Model architecture}\label{sec:network_architecture}
\subsubsection{Single-task learning}\label{single-task-learning}
In this paper, we refer to single-task learning as a setting where we train one separate model for each of the libraries. The terminology is used to contrast with multi-task learning where we train one unified model for all the editors/datasets. For the single-task learning, we used the two-stage model ( Figure \ref{fig:Two-step-Model}) with protospacer and PAM as the reference sequence representation.  In this section, we provide an in-depth introduction to the two-stage model architecture, which is comprised of two distinct sub-models: the overall efficiency model and the proportion model.

\paragraph{Overall Efficiency Model}
The Overall Efficiency Model concentrates exclusively on the target sequence, overlooking the specific edit outcomes. Its main objective is to predict the probability of the target sequence undergoing modification, regardless of the nature of the edits. Hence, the model exclusively processes the input target sequence, which in our scenario is the concatenation of the protospacer and PAM, yielding the probability of the target sequence undergoing modification (yielding non-wild type outcomes). To achieve this, we propose to use a Convolutional Neural Network (CNN) on the one-hot encoding of the target sequence. More specifically, we use three layers of 1D-CNN (kernel size: 2, stride: 2) with filter sizes of 32, 64, and 128, respectively. Following the CNN layers, we apply a feed-forward network with ReLU activation, featuring a hidden layer dimension of 64. The output of this network is a two-dimensional vector, which is subsequently transformed into probabilities through the use of the Softmax function.

\paragraph{Proportion Model}
Different from the absolute efficiency model, the proportion model focuses on predicting the probability of different types of edited outcomes for the target sequence. Therefore, it takes both the target sequence as well as one of its corresponding outcome sequences and outputs the probability of observing such an outcome. To implement this model, we use two encoder networks and one prediction network. The architectures of the two encoding networks, one for the target sequence and the other for the outcome sequence are identical, as illustrated in Figure \ref{fig:Two-step-Model}. Consequently, we will only describe in detail one of these networks here for clarity.
The target/reference sequence encoder network comprises two essential components: an embedding block and an encoder block.

\paragraph{Embedding Layer}
The embedding block embeds both the nucleotides and their corresponding position (in the protospacer) from the one-hot encoded representation to a dense vector representation. Given a protospacer sequence extended with its corresponding PAM site: $\bx_{\text{ref}} = [x_1, x_2, \dots,x_T]\in \mR^{T}$, $x_t$ represents the nucleotide at position $t$. In our case, T=24. We use $\mathbf O=[\mathbf o_1, \mathbf o_2, \dots, \mathbf o_T]\in \mR^{K\times T}$ as its one-hot encoded representation.  Here $K = 4$ as we have only four distinct nucleotides.  

An embedding matrix $\mathbf W_e$ is used to map each $\mathbf o_t\in \mR^k$ to a fixed-length vector representation:
\begin{equation}
    \mathbf e_t = \mathbf W_e\mathbf o_t
\end{equation}
where $\mathbf W_e\in \mR^{d_2\times K}$, $\mathbf e_t \in \mR^{d_e}$,  and $d_e$ is the embedding dimension we chose. 

Similarly, each nucleotide's position in the sequence $\bx_{\text{ref}}$ is represented by one-hot encoding with dictionary size $T$. En embedding matrix $\mathbf W_{p'}\in\mR^{d_e\times T}$ is applied to project the  $\mathbf p_4$ to a dense vector representation:
\begin{equation}
     \mathbf p'_t = \mathbf W_{p'}\mathbf p_t
\end{equation}
 where $\mathbf W_{p'}\in \mR^{d_2\times T}$, $\mathbf p_t \in \mR^{d_e}$. 
 Both embeddings $\mathbf e_t$ and $\mathbf p'_t$ are summed to get a unified representation for every element $x_t$ in the reference sequence $\bx_{\text{ref}}$. 
 \begin{equation}
     \mathbf u_t = \mathbf e_t + \mathbf p'_t \:\: \:\: \forall \:\:  t= 1, 2, \dots T
 \end{equation}
 This results in the embedded representation $\mathbf U=[\mathbf u_1, \mathbf u_2, \dots, \mathbf u_T]$ of the reference sequence. 

\paragraph{Encoder Block}
To learn a good representation that takes into account the relationships between the nucleotides in the reference sequence, we use a multi-head self-attention to encode the embedded representation. Multi-head Attention is a module that employs multiple single-head self-attention in parallel (i.e. simultaneously) to process each input vector $\mathbf u_t$. The outputs from every single-head layer are then concatenated
and transformed by an affine transformation to generate a fixed-length vector.

The single-head self-attention approach \citep{vaswani2017attention} learns three different linear projections of the input vector using  three separate  matrices: (1)
a queries matrix $\mathbf W_{query}$, (2) keys matrix $\mathbf W_{key}$, and (3) values matrix $\mathbf W_{value}$. Each input $\mathbf u_t$ in $\mathbf U$ is mapped
using these matrices to compute three new vectors:
\begin{align}
   & \mathbf q_t = \mathbf W_{query}\mathbf u_t\\
    &\mathbf k_4 = \mathbf W_{key}\mathbf u_t\\
    &\mathbf v_t = \mathbf W_{value}\mathbf u_t\\
\end{align}
where $\mathbf W_{query}$, $\mathbf W_{key}$, $\mathbf W_{value}\in \mR^{d\times d_e}$, $\mathbf q_t, \mathbf k_t, \mathbf v_t \in \mR^{d}$ are query, key and value vectors, and $d$ is the dimension of the those projected vectors. In the second step, attention scores are computed using the pairwise similarity between the query and key vectors for each position $t$ in the sequence. The similarity is defined by
first computing a scaled dot product between the pairwise vectors and then normalizing it using the softmax function. At each position $t$, we compute attention scores
$\alpha_{tl}$ representing the similarity between $t$-th query $\mathbf q_t$ and $l$-th key $\mathbf k_l$.
\begin{equation}
    score(\mathbf q_t, \mathbf k_l) = \frac{\mathbf q_t^T\mathbf k_l}{\sqrt{d}}
\end{equation}
\begin{equation}
    \alpha_{tl} = \frac{\exp(score(\mathbf q_t, \mathbf k_l))}{\sum_{l=1}^T \exp(score(\mathbf q_t, \mathbf k_l))}
\end{equation}
Then a weighted sum of value vector $\mathbf v_l$ using attention $\alpha_{tl}, \:\: \forall \: l\in \{1,2,\dots, T\}$ is performed to generate a new vector representation $\mathbf e_t\in \mR^d$ at position $t$. 

\begin{equation}
    \mathbf e_t = \sum_{l=1}^T \alpha_{tl}\mathbf v_l
\end{equation}
This process is applied to every position in the original embedding of the sequence, $\mathbf U$, to obtain a sequence of vectors $\mathbf E = [\mathbf e_1, \mathbf e_2, \dots, \mathbf e_T]$.

In a multi-head setting with H number of heads, the queries, keys, and values matrices will be indexed by superscript $h$ (i.e. $\mathbf W_{query}^h$, $\mathbf W_{key}^h$, $\mathbf W_{value}^h\in \mR^{d\times d_e}$) and applied separately to generate a new vector representation $\mathbf e_t^h$ for every singe-head self-attention layer. The output from each single-head attention layer is contenated into one vector $\mathbf e_t^{concat}=concat(\mathbf e_t^1, \mathbf e_t^2, \dots, \mathbf e_t^H)$ where $\mathbf e_t^{concat}\in \mR^{dH}$. Then it goes through an affine transformation using $\mathbf W\in \mR^{d\times dH}$ and $\mathbf b\in \mR^d$ to generate the encoded representation $\hat{\mathbf{Z}} = [\hat{\bz}_1, \hat{\bz}_2, \dots,\hat{\bz}_T]$ of the reference sequence
\begin{equation}
    \hat{\bz}_t = \mathbf W\mathbf e_t^{concat} + \mathbf b
\end{equation}

To improve the gradient flow in layers during
training, we also use residual connections / skip-connections \citep{he2016deep}. This is done by summing both the newly computed output of the current layer with the output from the
previous layer. In our setting, a first residual connection sums the output of the self-attention layer $\hat{\bz}_t$ and the
output of embedding block $\mathbf u_t$ for each position $t$ in the sequence.

We also deploy layer normalization \citep{ba2016layer} after the self-attention layer 
with the goal of ameliorating the "covariate-shift" problem by re-standardizing the computed vector representations
(i.e. using the mean and variance across the features/embedding dimension $d$). Given a computed
vector $\hat{\bz}_t$, the LayerNorm function will standardize the input vector using the mean and variance
along the dimension of the feature $d$ and apply scaling 
 and shifting steps.

Eventually, this learned representation goes through a feedforward network with one hidden layer and ReLu activation function. Subsequently, a layer normalization is applied to the output of this feed-forward network to obtain the learned representation $\bz_t\in \mR^d$. Eventually, the encoder block transformed the embedded vector $U$ to the learned representation $\mathbf Z=[\bz_1, \bz_2, \dots, \bz_T]\in \mR^{T\times d}$ that incorporates the contextual information/relationships between features through attention.

The above process describes one encoding block, we stack N such blocks to construct our encoder network. As it is presented in Figure \ref{fig:Two-step-Model}, in the proportion model, we apply two encoder networks with the same network architecture on the reference sequence and corresponding outcome sequence. This yields two encoded representations which we denote by $\mathbf Z^{ref}\in \mR^{T\times d} $ and $\mathbf Z^{out}\in \mR^{T\times d} $ respectively. We then concatenated them in the feature dimension to generate one common representation $\mathbf Z= [\bz_1, \bz_2, \dots, \bz_T]\in \mR^{T\times 2d}$. 

\paragraph{Output network} 
The output network consists of an affine transformation and a nonlinear function to transform the output to probability:

\begin{equation}
    \by_{out,t} = \mathbf W_{out}\bz_t
\end{equation}
where $\mathbf W_{out}\in \mR^{2d\times 2}$,$\by_{out,t}\in \mR^2$. We then apply a softmax function on $\by_{out,t}$ and transform it to a probability that represents the probability of the nucleotide at position $t$ getting edited. Note that, we use the same length of input and outcome sequence. However, our input sequence also includes PAM information, and editing only happens in the protospacer. Moreover, due to the nature of the Base editor, only specific nucleotides, in our case Adenine (A) gets edited while the other nucleotides are not affected. Therefore, we use masking technique to only consider the positions that are possible to be edited and mask out other positions. Therefore, the PAM information (or the contextual information such as left/right overhangs is participating by affecting the embedding of the nucleotides in the protospacer but is not considered in the loss as they are not changed/edited.

After tuning the parameters, for the Proportion Model, we have chosen an embedding dimension of 124 for the embedding layer. Our model consists of 12 encoding blocks, with each block featuring 8 multi-head attention mechanisms. For the output network, we employ a single linear layer that maps a 248-dimensional vector to a two-dimensional output vector. Subsequently, we transform this output into probabilities using the softmax function.

\paragraph{Multi-task Learning }
We extended the two-stage model for accommodating various base editors through the implementation of a multi-task learning framework, eliminating the need for training individual models per base editor.

To achieve this, we first augment the datasets from different libraries with corresponding editor labels and combine all the libraries to create a consolidated dataset. Our objective is to establish a shared architecture comprising common layers applicable across all libraries, along with dedicated sub-networks tailored to each specific library. 

As illustrated in Figure \ref{fig:multi_task}, we begin by extending the overall efficiency model by incorporating the first two layers of a 1D-CNN as the universally shared layers for all libraries. Subsequently, this shared learned representation traverses a sub-network comprising two layers of 1D-CNN and two layers of MLP with ReLU activation functions, uniquely customized for each library. Given that, as we have datasets from six editors, we use one common network (consisting of two layers of CNN)  and six distinct sub-networks, which have identical structures. 

For the proportion model, the influence of various libraries/Base Editors is observable through the variations in the outcome set corresponding to the various base editors when applied to the same target sequence. Therefore, to extend the proportional model for various base editors, we maintain a shared encoder network for the reference sequence across all libraries while constructing six distinct encoder networks to encode the outcome sequences from the six different libraries. This approach allows us to establish a consistent representation for the reference sequences across all libraries, while simultaneously accommodating the distinctions among the libraries by employing separate encoder networks to encode the outcome sequences for each one. Consequently, our multi-task learning proportional model comprises one shared reference sequence encoder network, six individual outcome sequence networks, and six corresponding output networks.
\subsection{Optimization}
For optimization, we use Adam optimizer \citep{kingma2014adam} with a learning rate scheduler. We initialize the base learning rate at $3e^{-4}$ and set the maximum learning rate to five times the base rate.  Additionally, we incorporate dropout \citep{srivastava2014dropout} probability of 0.2. The regularizer parameter $\lambda $ is set to $1e^{-4}$. 

It's worth noting that while it's possible to train both the overall efficiency model and the proportion Model simultaneously. However, 
Training them together means, we apply the loss on the learned final probability,  $\hat{P}(X_{out}|\bx_{ref})$ (Eq. \ref{model:full}), with the true probability which represents the probability of all outcomes. This means that for the edited (non-wild outcome) outcomes, the problem of true probability being very low compared to the wild-type outcomes still exists. Breaking down this final probability into a product of two conditional probabilities was introduced to mitigate such problem as $P(X_{out}|\bx_{ref}, edited)$ focuses only on the non-wild type outcomes. Therefore, training them together could result in an outcome where the model predominantly focuses on predicting easily predictable outcomes (as wild type) while neglecting those with lower probabilities as we hypothesize in the One-stage model. Moreover, the two models do not share any common layer except the outcome of the two networks gets multiplied to generate the final probability distribution. Therefore, there is no real requirement by model design to train them together. To avoid falling back to the single-step model, we train the Absolute Efficiency Model and the Proportion Model separately. This approach explicitly matches the probabilities $P(C|\bx_{\text{ref}})$ and $P(X_{out}|\bx_{\text{ref}}, \text{edited})$, preventing the model from overlooking low-probability outcomes.

In terms of training specifics, we set the mini-batch size to 100 and the maximum number of epochs to 300 for the Absolute Efficiency Model. For the Proportion Model, we choose a mini-batch size of 400 and an epoch count of 150. In both models, we use Spearman correlation on the validation set as our performance metric to monitor and select the best-performing model.

\subsection{Performance measures} 
Considering the distinctive nature of our data generation process, we chose to employ Pearson and Spearman correlations as our performance metrics, measuring the alignment between actual and predicted probability scores. Owing to inherent variability during screening, repeated experiments under identical conditions yield slightly divergent outcomes. For the biologist, the emphasis is on correlating these similar results, rendering metrics such as mean square or mean absolute error less pertinent. Our primary concern isn't exact prediction precision, but rather the level of correlation achieved between predictions and actual data.

\subsection{Comparing with the other baselines}\label{sec:baseline_comparision}
In addition to table \ref{tab:baseline_2}, to further fairness,  we also assessed our two-stage model, which includes a sub-model (the proportion model) trained exclusively on non-wild-type outcomes, akin to the baseline models, using the dataset from \cite{song2020sequence}. This dataset includes PAM information in its reference sequence. However, it's worth noting that the PAM in their dataset consists of only three bases, while our model is trained on PAMs with four bases. 

To be able to use our model, we extended one of the neighboring bases to serve as the fourth PAM base. Note that our model by design is not restricted to having four bases for PAM.  It's worth noting that our model is inherently flexible and not constrained to a fixed PAM size of four bases. The ideal approach would involve retraining our model on the training set from \cite{song2020sequence} and reporting the results. As an initial exploration, we applied our existing trained model to assess its performance, even with a mismatched base for the last PAM position. The results, presented in Table \ref{tab:base_line_3}, indicate a slight performance decrease but still demonstrate competitive performance relative to most of the baseline models.
\begin{table}
   \centering
    \begin{tabular}{lll}
    \toprule
         &overall &non-wild type \\
         \hline
       Our two-stage model  & 0.923& 0.942\\
       \hline
    \end{tabular}
    \caption{Our model performance in terms of Pearson correlation on the \cite{song2020sequence} testset}
   \label{tab:base_line_3}
\end{table}
Finally, we also tested our model with respect to the most recent model DeepBE\cite{kim2023deep}. We have trained our model on one of the datasets they provide (randomly chosen) and compared our result with the ones reported in their paper. The chosen dataset corresponds to the results from the ABE8e-NG base editor. This data includes 7112 unique target sequences and overall 28261 input-output pairs. The data they provide is processed such that only bases from position 3 to 10 (editing window size of 8) are considered to be edited. The data has train/test separation where 6340 among 7112 target sequences are selected as training and the rest as a test set. We trained our model on their training set and tested our model performance on their test set. Here is the performance comparison is given in Table \ref{tab:baseline_DeepBE}.
\begin{table}[]
\centering
    \resizebox{0.7\linewidth}{!}{%
    \begin{tabular}{lll}
    
\hline
Model & Pearson correlation & Spearman correlation \\
\hline
DeepBE \cite{kim2023deep} & 0.83 & 0.91 \\
Ours & 0.89 & 0.96 \\
\hline
\end{tabular}}
\caption{\small{Comparing with DeepBE peformance on one of their dataset corresponding to ABE8e-NG base editor}}
    \label{tab:baseline_DeepBE}
\end{table}
\subsubsection{Multi-task learning}
 To extend our two-stage model in the setting of multi-task learning (\ref{sec:Multi_task_learning}), we explored two distinct methodologies for tackling multi-task learning. The first involves a direct conversion of the distribution into a conditional form, conditioned upon the editor label. The second applies a structural transformation of the network enabling the model to have both shared and distinct layers across various libraries. We refer to the first as a conditional model and the second as multi-task learning. 
 In order to identify a suitable approach, we exclusively assessed both methodologies using the absolute efficiency model, leveraging its inherent simplicity. This choice stems from the rationale that if the conditioning factor is overlooked within this inherently simpler context, its impact is likely to be minimal when applied to the proportion model, which is considerably more intricate. As illustrated in Table \ref{tab:conditional_model}, the multi-task setup on the absolute efficiency model has a substantial advantage over the model that uses the editor label as a conditioning factor.

\begin{table}[H]
    \centering
      \resizebox{\linewidth}{!}{%
     \begin{tabular}{lllll}
     \toprule
     & \multicolumn{2}{c}{ Conditional model}& \multicolumn{2}{c}{Multi task learning}\\
       Libraries & Spearman& Pearson&Spearman&Pearson\\  
          \hline
     SpRY-ABEmax & $0.677 \pm 0.004$ & $0.629 \pm 0.003$ & $0.797 \pm 0.007$ & $0.783 \pm 0.012$ \\
    SpCas9-ABEmax & $0.811 \pm 0.009$ & $0.759 \pm 0.002$ & $0.834 \pm 0.011$ & $0.901 \pm 0.002$ \\
     SpG-ABEmax & $0.811 \pm 0.009$ & $0.748 \pm 0.006$ & $0.853 \pm 0.009$ & $0.835 \pm 0.017$ \\
     SpRY-ABE8e & $0.548 \pm 0.012$ & $0.537 \pm 0.018$ & $0.578 \pm 0.031$ & $0.695 \pm 0.034$ \\
     SpCas9-ABE8e & $0.751 \pm 0.010$ & $0.723 \pm 0.016$ & $0.866 \pm 0.031$ & $0.862 \pm 0.014$ \\
     SpG-ABE8e & $0.788 \pm 0.005$ & $0.760 \pm 0.011$ & $0.788 \pm 0.002$ & $0.824 \pm 0.004$ \\
     \bottomrule
     \end{tabular}}
     \caption{\small{Performance comparison of overall efficiency model on two different settings: conditional model $P(C|\bx_{\text{ref}}, B_i)$ and multi-task learning approach }}
\label{tab:conditional_model}
\end{table}

\subsection{More results}
To gain deeper insights into the model's performance, we provide scatter plots showcasing the actual and predicted probability values derived from the multi-task model on the SpRY-ABEmax library test set. It's important to note that the library selection is entirely random, and this particular library is not cherry-picked; similar results are observed across all other libraries as well. 
\begin{figure}[tb]
    \centering
    \includegraphics[scale=0.5]{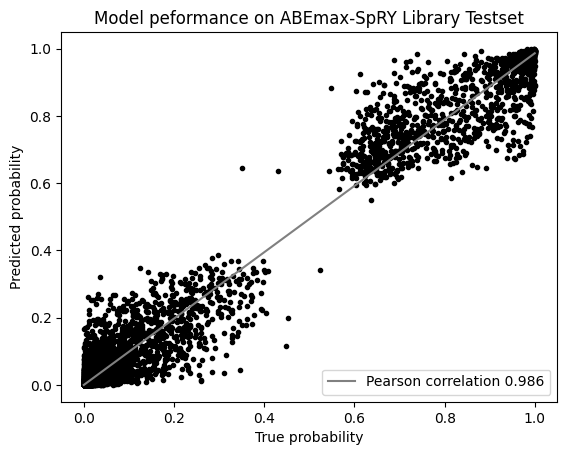}
    \caption{Performance of the Multi-Task Model Across All Possible Outcomes, Including Both Wild-Type and Non-Wild-Type Variants}
    \label{fig:scater1}
\end{figure}
\begin{figure}[tb]
    \centering
    \includegraphics[scale=0.5]{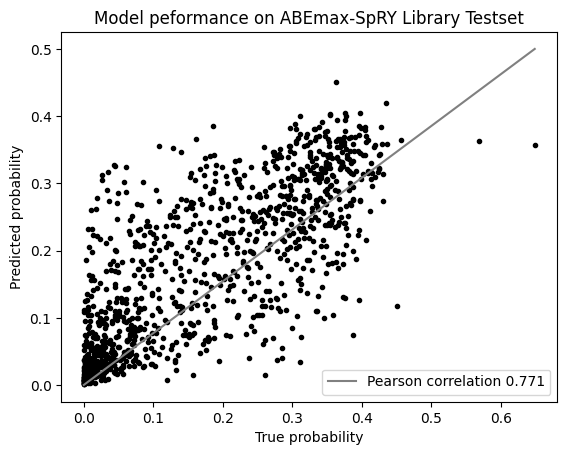}
    \caption{Performance of the Multi-Task Model Across  Wild Type (this corresponding to the model absolute efficiency model performance)}
    \label{fig:scater2}
\end{figure}
\begin{figure}[H]
    \centering
    \includegraphics[scale=0.5]{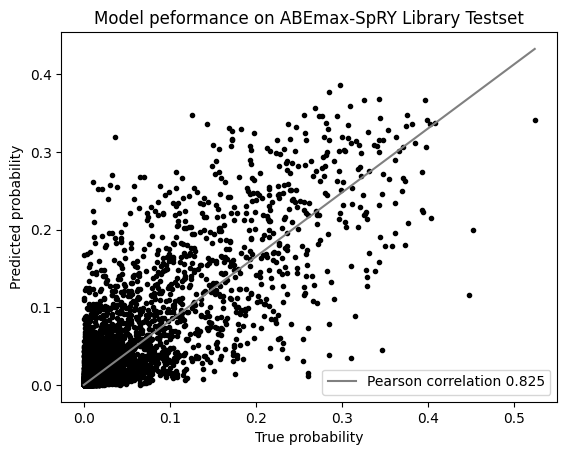}
    \caption{Performance of the Multi-Task Model Across Non-Wild Type}
    \label{fig:scater3}
\end{figure}
\begin{figure}[H]
    \centering
    \includegraphics[scale=0.5]{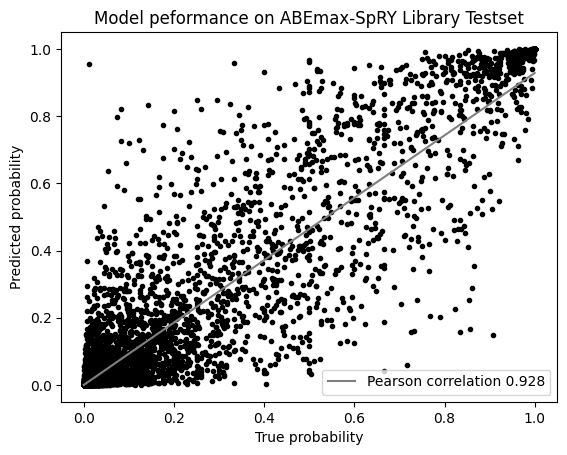}
    \caption{Performance of the Multi-Task Model Across Non-Wild Type (presented in the randomized space, i.e., $P(X_{out}|\bx_{\text{ref}, edited}$, this corresponds to the proportional model performance)}
    \label{fig:scater4}
\end{figure}
\newpage

\end{document}